# A 21-DOF Humanoid Dexterous Hand with Hybrid SMA-Motor Actuation: CYJ Hand-0 *

## Jin Chai, Xiang Yao, Mengfan Hou, Yanghong Li and Erbao Dong*

*Abstract*—CYJ Hand-0 is a 21-DOF humanoid dexterous hand featuring a hybrid tendon-driven actuation system that combines shape memory alloys (SMAs) and DC motors. The hand employs high-strength fishing line as artificial tendons and uses a fully 3D-printed AlSi10Mg metal frame designed to replicate the skeletal and tendon-muscle structure of the human hand. A linear motor-driven module controls finger flexion, while an SMA-based module enables finger extension and lateral abduction. These modules are integrated into a compact hybrid actuation unit mounted on a custom rear support structure. Mechanical and kinematic experiments, conducted under an Arduino Mega 2560-based control system, validate the effectiveness of the design and demonstrate its biomimetic dexterity.

*Index Terms*-- Humanoid dexterous hand, hybrid actuation, tendon-driven mechanism, 3D printing, biomimetic robotics

## I. INTRODUCTION

From the late 20th century to the early 21st century, the concept of "robots" has transitioned from science fiction into tangible reality. Initially designed for simple, specialized tasks, robots have progressively undertaken increasingly complex, precise, and critical roles in industry and military. Among these, robotic manipulators stand as the earliest industrial robots and the foundation of modern robotics, liberating humans from laborious tasks and facilitating mechanization and automation. Moreover, they operate effectively in hazardous environments, thereby enhancing safety in sectors including machinery manufacturing, metallurgy, electronics, light industry, and nuclear energy.

The human hand represents the pinnacle of biological evolution, exhibiting extraordinary dexterity and versatility. Humanoid dexterous hands (HDHs) strive to replicate the intricate form, structure, and function of the human hand to enable similarly flexible and delicate manipulation. They hold immense promise for diverse applications spanning medical assistance, domestic services, intelligent manufacturing, space exploration, oceanic development, and public safety [1].

The history of prosthetic hands dates back to Roman times circa 200 B.C., where early wooden prostheses primarily served cosmetic or rudimentary functional purposes. The 19th century witnessed the invention of the tendon-driven, single-degree-of-freedom prosthetic hand by Peter Baliff, powered by the user's limb movement—a fundamental principle that persists in contemporary prosthetics. The advent of the Belgrade hand [2] by Tomovic and Bonic in 1962 marked a milestone, ushering in a new era of HDHs research. Since then, academic institutions and technology companies worldwide have pursued extensive research, yielding significant advancements such as the Okada hand [3], Stanford/JPL hand [4], Utah/MIT hand [5], and Robonaut hand [6]. With the advent of the 21st century, breakthroughs in sensor technology, human-computer interaction, bionics, and artificial intelligence have driven the evolution of HDHs towards greater diversity, miniaturization, and intelligence, exemplified by innovations like the Cyber hand[7], Tokyo hand [7] and UT hand [8].

Despite these advances, existing HDHs are often constrained by limited structural biomimicry, insufficient degrees of freedom (DOF), inadequate sensory capabilities, limited flexibility and coordination, low load-bearing capacity, fragile impact resistance, and high manufacturing costs. To address these challenges, the CYJ Hand-0 proposed herein incorporates the following key features:

- A highly biomimetic skeletal and joint structure with 21 DOFs (excluding the wrist).
- A flexible tendon-driven system utilizing high-strength fishing line.
- A hybrid actuation scheme integrating a shape memory alloy (SMA) linear drive module with double-amplified stroke and a DC brush motor linear drive module.
- Integration of the actuation system into a custom-designed robotic arm.
- Fabrication through cost-effective, easily manufacturable AlSi10Mg metal 3D printing, resulting in a total mass of only 380 grams.
- A fundamental control system based on the Arduino Mega2560 microcontroller for single-finger and full-hand operation.
- Single-finger load capacity of 1.2 kgf and an overall hand load capacity of 8 kgf.
- Capability to perform all Kapandji tests, 32 distinct gesture actions, and over 30 grasping experiments.

The remainder of this paper is structured as follows: The first section elaborates on the hand's overall structural design, key components, kinematic modeling, and mechanical analysis of the flexible tendon drive. The second section details the SMA linear flexible drive module, DC brush motor linear drive module, system integration, and control framework. The last section presents performance evaluations in force, motion, dexterity, and grasping ability. Finally, the Conclusion summarizes the work and discusses future research directions.

## II. MATERIALS AND METHODS

### A. Bionic principle

*Research supported by the National Science Foundation of China (Grant number: 61773358).

All authors are associated with the key Laboratory of Precision and Intelligent Chemistry, Institute of Humanoid Robots, Department of Precision Machinery and Precision Instrumentation, University of Science and Technology of China, Hefei, Anhui, 230026, PRC (*Corresponding author is Erbao Dong and can be contacted at e-mail: ebdong@ustc.edu.cn)

The human hand's main anatomical structures can be categorized into eight primary components: bones, tendons, ligaments, muscles, blood vessels, nerves, fatty tissue, and skin. The skeletal system is notably complex, comprising a total of 27 bones, which include 8 carpal bones, 5 metacarpal bones, and 14 phalanges [9]. The carpal bones are located within the wrist and are classified as radial or ulnar based on their anatomical position. The metacarpal bones articulate with the carpal bones at their bases via the carpometacarpal (CMC) joints, while the phalanges of each finger connect to the metacarpophalangeal (MCP) joints proximally. For the four fingers other than the thumb, each joint chain consists of the MCP joint, proximal interphalangeal (PIP) joint, and distal interphalangeal (DIP) joint, progressing from the proximal to the distal end. Due to mechanical constraints inherent to the interphalangeal joints, the MCP joint allows flexion and extension between 0° and 90°, and lateral deviation between −15° and 15°. The PIP joint exhibits a flexion range from 0° to 110°, while the DIP joint allows motion from 0° to 90° [10].

Furthermore, based on anatomical principles, the DIP joint flexes simultaneously with the PIP joint in the index, middle, ring, and little fingers, establishing a dynamic intra-finger coupling. This relationship is mathematically expressed in Eq. (1) and illustrated in Fig. 1.

$$\theta_{\text{DIP}} = k\theta_{\text{PIP}}, \ k \approx \frac{2}{3} \tag{1}$$

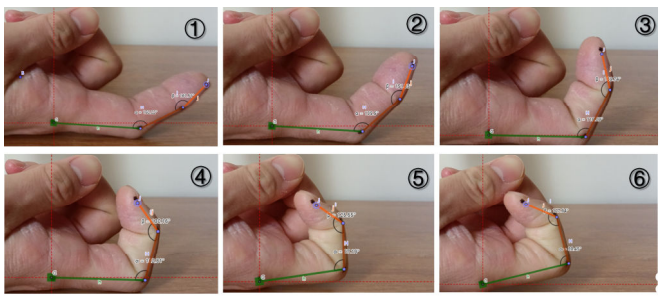

**Figure 1. Angle changes between knuckles during index finger bending.**

The structure of the thumb differs from that of the other four fingers in that it comprises only three joints: the carpometacarpal (CMC), metacarpophalangeal (MCP), and distal interphalangeal (DIP) joints, lacking a proximal interphalangeal joint. The thumb's carpometacarpal joint, also known as the trapeziometacarpal (TMC) joint, grants it significantly greater mobility and strength compared to the other fingers. This joint facilitates a wide range of motions, including rotation (~17°), adduction and abduction (~42°), flexion and extension (~45°), as well as opposition—the ability to cross the palm. Consequently, the thumb exhibits exceptional dexterity. Traditionally, the TMC joint was considered to have 3-DOF [11], which, combined with the rest of the hand, results in a total of 21-DOF. However, subsequent research has revealed that the articular surfaces between the trapezium and the first metacarpal are saddle-shaped, with the axes for adduction-abduction (Ab-Ad) and flexion-extension (Fl-Ex) being neither orthogonal nor intersecting [12]. Most contemporary biomimetic hands, such as the Etho hand, implement the thumb's TMC joint as a 2-DOF ball joint. While ball-and-socket joints provide smooth movement that approximates 2-DOF, their structural configuration differs from that of the human thumb and typically requires up to three actuators and six tendons for control.

The DOFs of the human hand are commonly simplified to 20 by omitting subtle motions, including active or passive rotation of each finger around its longitudinal axis [13] (as illustrated in Fig. 2a), the complex kinematics of the thumb's

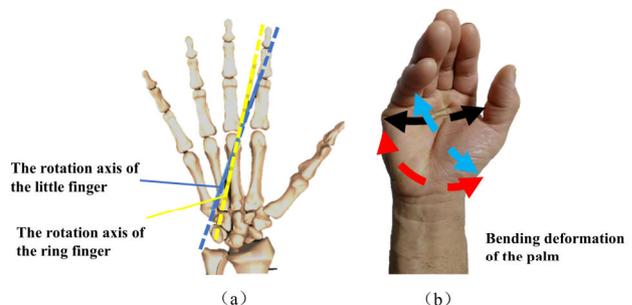

**Figure 1. The rotation axes of the fingers [13] and palm flexion.**

TMC joint, flexion and deformation of the palm (Fig. 2b), and the DOFs of the wrist.

### B. Overall structure and 3D prototype

The overall structural design of the CYJ Hand-0 is illustrated in Fig. 3a. All components are named according to their anatomical counterparts in the human hand, including the thumb, index finger, middle finger, ring finger, little finger, metacarpals, carpals, and wrist. Notably, the thumb's metacarpal structure is divided into two distinct segments: the medial abduction carpometacarpal joint and the rotational carpometacarpal joint, which will be described in detail later. In addition, the index, middle, and ring fingers adopt a unified structural design to simplify fabrication and enhance component interchangeability.

As shown in Fig. 3b, the CYJ Hand-0 consists of 18 distinct component types, totaling 80 individual parts. To improve modularity and reduce manufacturing costs, many joints share identical components. The hand is not only anatomically inspired in appearance but also replicates the skeletal structure of a human hand at a 1:1 scale in both form and assembly, as depicted in Fig. 3c. The total mass of the hand is 380 g, which accounts for less than 0.575% of the body weight of an average 75 kg adult male (approximately 413 g for a human hand). As shown in Fig. 3d, the unique mechanical design of the bionic hand enables up to 21 DOFs, excluding the wrist.

### C. Structural design of key components

The index finger structure of the CYJ Hand-0 serves as a universal design, also applied to the middle and ring fingers. Each universal finger consists of 9 distinct component types—including the proximal, middle, and distal phalanges—resulting in a total of 12 individual parts. To ensure optimal functional performance and structural integrity, the design incorporates sliding bearings and adopts a biomimetic shape that closely replicates the interphalangeal joints of the human finger. The structural layout and dimensions of the universal finger are shown in Fig. 4. The little finger differs from this universal structure only in the length of the proximal phalanx and a few minor structural details, as illustrated in Fig. 5a.

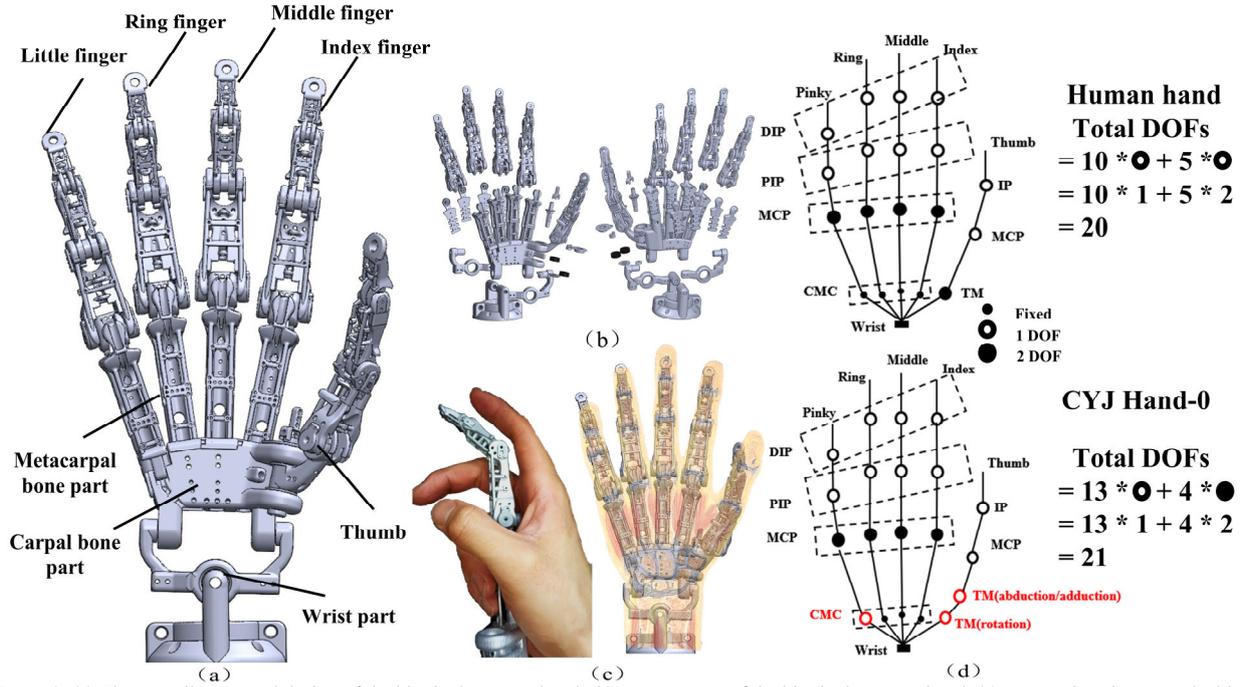

**Figure 3.** (a) The overall structural design of the bionic dexterous hand. (b) Components of the bionic dexterous hand. (c) Comparison between the bionic dexterous hand and human hand. (d) Comparison of DOFs between the bionic dexterous hand and human hand.

The thumb, being the most flexible and complex digit of the CYJ Hand-0, adopts a customized structural design. To balance structural simplicity with functional fidelity, the thumb's metacarpal is divided into two segments at the carpometacarpal (CMC) joint: one responsible for adduction/abduction and the other for flexion/extension. The adduction/abduction module is structurally similar to the metacarpophalangeal (MCP) joints of the other fingers. This design choice reduces the overall degrees of freedom, enhances joint modularity, and simplifies mechanical implementation, as shown in Fig. 5b.

The joint rotation ranges of the CYJ Hand-0 are designed to match those of the human hand and are compared with representative existing bionic hands in Table I.

TABLE I. COMPARISONS OF THE HDHS

| Hand | Weight (g) | Interphalangeal joint rotation angle | | | Thumb joint rotation angle | |
|---|---|---|---|---|---|---|
| | | *MCP* (°) | *PIP* (°) | *DIP* (°) | *Ab-Ad* (°) | *Ex-Fl* (°) |
| Human hand | 431 | 0-90 | 0-110 | 0-90 | 0-70 | 0-70 |
| CYJ Hand-0 | 380 | 0-90 | 0-110 | 0-90 | 0-53 | 0-107 |
| Robotic hand | | 0-90 | 0-110 | 0-90 | 0-90 | 0-90 |
| Tact hand | 350 | 0-90 | 23-90 | 0-20 | 0-90 | 0-105 |
| Dextrus hand | 428 | 0-90 | 0-90 | 0-90 | 0-90 | 0-120 |
| Bebionic hand | 550 | 0-90 | 0-90 | 0-20 | | 0-68 |
| i-Limb hand | 600 | 0-90 | 0-90 | 0-20 | 0-60 | 0-95 |

The metacarpal bones of the human hand are numerous, tightly packed, and arranged in a complex three-dimensional configuration. To enhance the anatomical fidelity of the palmar structure in the CYJ Hand-0—while minimizing unnecessary internal space and reducing weight—an interchangeable metacarpal design and an integrated carpal bone structure were developed, as illustrated in Fig. 5c. Notably, the metacarpal bones of the four fingers and the base connecting to the carpal region are not arranged in a parallel manner. Instead, they are positioned at specific angles relative to one another, forming a naturally concave geometry that mimics the human palm. This curvature enables the CYJ Hand-0 to achieve a cupped palm configuration, thereby enhancing its grasping performance, as shown in Fig. 5d.

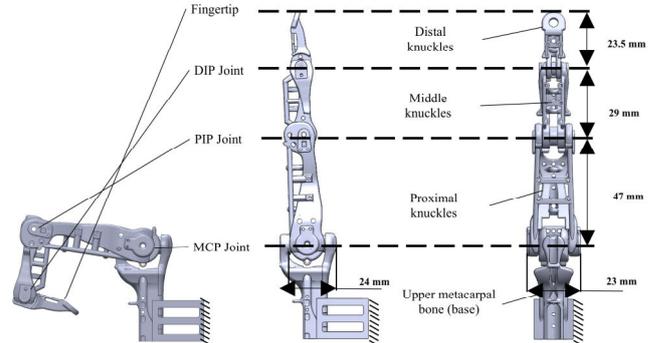

**Figure 4. Structure and size of the universal finger.**

### D. Kinematic modeling of the fingers

Each finger of the CYJ Hand-0 can be thought of as a small, open-chain, rotationally articulated, tandem robot. The entire hand is equivalent to multiple such robots connected in parallel. The universal finger mechanism belongs to the side-swinging type 4-joint finger structure, as shown in Fig. 6a. The axes of joints 1 and 2 at the metacarpophalangeal joint are orthogonal to each other. Joints 3 and 4 represent the proximal and distal phalanges respectively. The axes of joints 2, 3 and 4 are always parallel. the finger linkage coordinate system is established according to the D-H theory, as shown in Fig. 6b.

Define $x_0 y_0 z_0$ as the fixed reference coordinate system 0. Define the origin as the point where the axes of joints 1 and 2 intersect. $X$ and $Y$ form the right-handed system. The lengths of the phalanges are $a_1$, $a_2$, $a_3$ and $a_4$. The angles of rotation of the joints are $\theta_1$, $\theta_2$, $\theta_3$ and $\theta_4$. The axes of the lateral pendulum joints at the metacarpophalangeal joints are perpendicular to the axes of the flexion joints at angles $\alpha_1 = 90°$. The coordinates of the position of the finger tip in the fixed reference system are $(x, y, z)$. See Table II below for detailed D-H parameters.

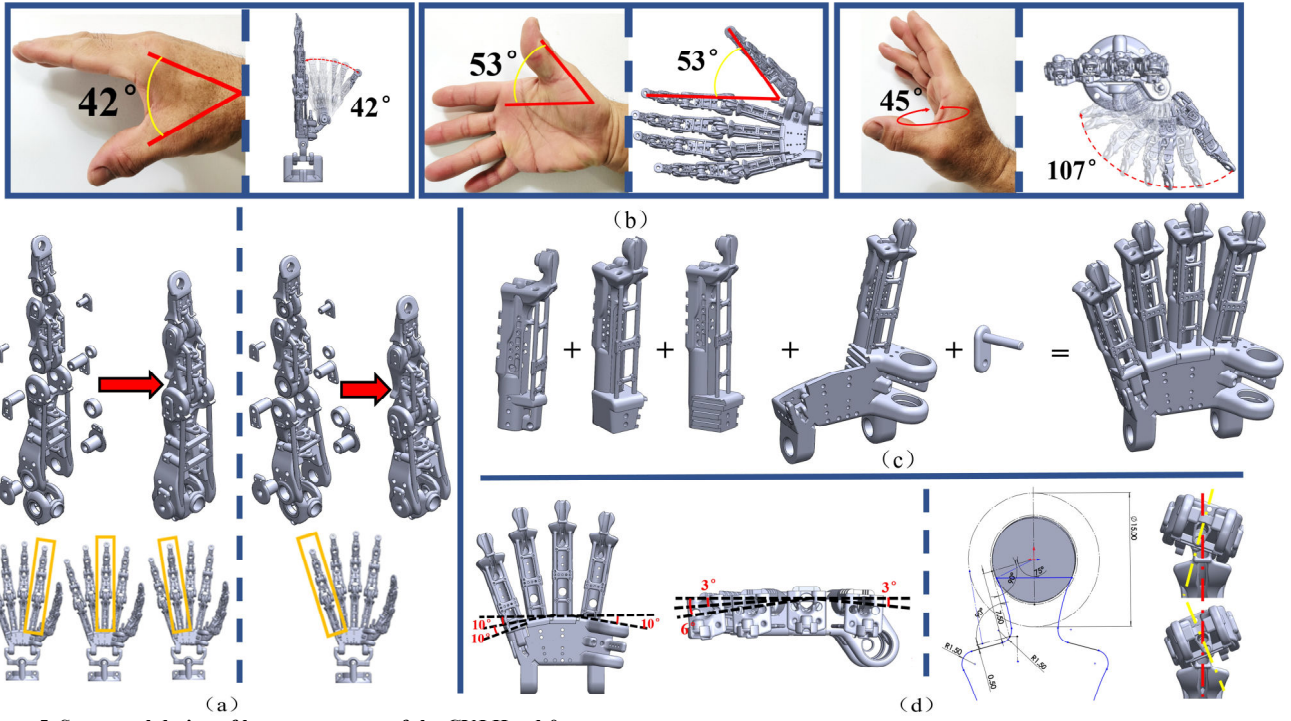

**Figure 5. Structural design of key components of the CYJ Hand-0.**

TABLE II. D-H PARAMETERS FOR THE UNIVERSAL FINGERS

| | $\alpha_{i-1}$/(°) | $a_i$/mm | $d_i$/mm | $\theta_i$/(°) |
|---|---|---|---|---|
| 1 | 0 | 0 | 0 | -15~15 |
| 2 | 90 | 47 | 0 | 0~90 |
| 3 | 0 | 29 | 0 | 0~110 |
| 4 | 0 | 23.5 | 0 | 0~90 |

The following paragraph discusses the solution to forward kinematics for a finger. The homogeneous transformation matrix that describes the position and orientation of the coordinate system at the fingertip relative to the fixed reference coordinate system 0 is:

$${}^{0}_{4}T = {}^{0}_{1}T \, {}^{1}_{2}T \, {}^{2}_{3}T \, {}^{3}_{4}T$$

$$= \begin{pmatrix} c_1 c_{234} & -c_1 s_{234} & s_1 & (a_4 c_{234} + a_3 c_{23} + a_2 c_2) c_1 \\ s_1 c_{234} & -s_1 s_{234} & -c_1 & (a_4 c_{234} + a_3 c_{23} + a_2 c_2) s_1 \\ s_{234} & c_{234} & 0 & -(a_4 s_{234} + a_3 s_{23} + a_2 s_2) \\ 0 & 0 & 0 & 1 \end{pmatrix}$$

(2)

In the Eq. (2), $s_i = sin\theta_i$, $c_i = cos\theta_i$, $s_{ij} = sin(\theta_i + \theta_j)$, $c_{ij} = cos(\theta_i + \theta_j)$, $s_{ijk} = sin(\theta_i + \theta_j + \theta_k)$, $c_{ijk} = cos(\theta_i + \theta_j + \theta_k)$, $i, j = 1, 2, 3, 4$.

Therefore, with known values of $\theta_1$, $\theta_2$, $\theta_3$ and $\theta_4$, the position coordinates $(x, y, z)$ of the general fingertip can be determined as follows:

$$x = [a_4 cos(\theta_2 + \theta_3 + \theta_4) + a_3 cos(\theta_2 + \theta_3) + a_2 cos\theta_2] cos\theta_1$$
$$= [23.5 cos(\theta_2 + \theta_3 + \theta_4) + 29 cos(\theta_2 + \theta_3) + 47 cos\theta_2] cos\theta_1$$
(3)

$$y = [a_4 cos(\theta_2 + \theta_3 + \theta_4) + a_3 cos(\theta_2 + \theta_3) + a_2 cos\theta_2] sin\theta_1$$
$$= [23.5 cos(\theta_2 + \theta_3 + \theta_4) + 29 cos(\theta_2 + \theta_3) + 47 cos\theta_2] sin\theta_1$$
(4)

$$z = -[a_4 sin(\theta_2 + \theta_3 + \theta_4) + a_3 sin(\theta_2 + \theta_3) + a_2 sin\theta_2]$$
$$= -[23.5 sin(\theta_2 + \theta_3 + \theta_4) + 29 sin(\theta_2 + \theta_3) + 47 sin\theta_2]$$
(5)

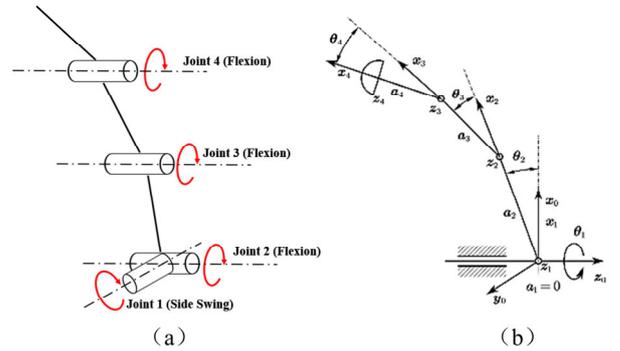

**Figure 6. The schematic diagram of the general finger mechanism and the D-H method coordinate system.**

The following text addresses the inverse kinematics solution for the finger. By substituting into Eq. (3) and (4), we obtain:

$$\theta_i = arctan \frac{y}{x}$$
(6)

By further substitution into Eq. (1), we obtain:

$$a_4 cos(\theta_2 + \theta_3 + \theta_4) + a_3 cos(\theta_2 + \theta_3) + a_2 cos\theta_2 = \frac{x}{cos\theta_i} = \frac{y}{sin\theta_i}$$
(7)

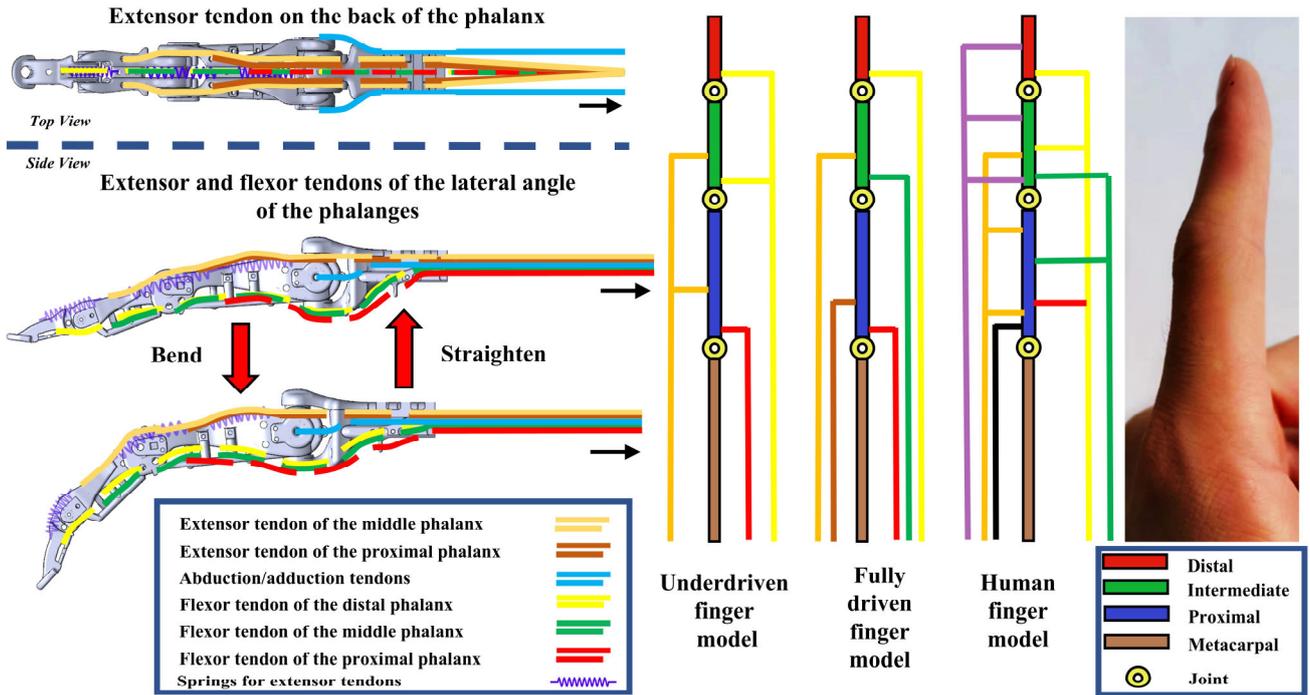

**Figure 7. Tendon distribution of the CYJ Hand-0.**

$$\frac{1}{a_2}cos\frac{2}{3}\theta_3 + \frac{1}{a_3}cos\frac{5}{3}\theta_3 + \frac{1}{a_4}cos\theta_3 = \frac{z^2 + t^2 - a_2^2 - a_3^2 - a_4^2}{2a_2a_3a_4} \quad (8)$$

The joint angles $\theta_3$ and $\theta_4$ can be reliably solved through numerical methods, and then substituted back into Eq. (3) to solve for $\theta_2$ using Newton's iteration method. Thus, with the known position coordinates of the fingertip ((x, y, z)), the joint angles $\theta_1$, $\theta_2$, $\theta_3$ and $\theta_4$ can be determined.

### E. Numerical simulation for finger tendon

The CYJ Hand-0 utilizes high-powered fishing line as a tendon. These tendons are fixed at one end to the phalanges and connected to linear actuators at the other, transmitting force and torque over a distance. Such tendons possess high strength, light weight, and an ability to absorb impacts that gear transmissions lack. Here we use a 12-braid high-powered fishing line with a mark of 9.0, a diameter of 0.55mm and a nominal maximum pull of 40kg. At the distal end of the tendons, the rope is commonly secured using the figure-eight loop method and the double fisherman's knot technique.

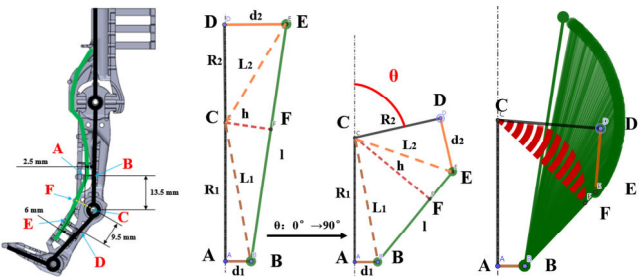

**Figure 8. Geometric analysis of tendons.**

The tendons have a fixed path of travel at each finger joint. Grease is used to reduce friction at the edges of holes, tubes and wire structures where the rope rubs against the alloy components. As shown in Fig. 7, the universal finger is driven by seven tendons (excluding springs). The flexor tendons are located on the palmar side of the finger phalanges and there are three of them that drive the proximal, middle and distal phalanges in flexion. The extensor tendons, located on the dorsal side of the finger phalanges, are two in number and drive the proximal and middle phalanges to extend back to their original positions. The distal phalanx is not driven by an additional tendon and is reset by a pre-tensioned spring. The two abduction/adduction lateral tendons, located on both sides of the finger, with one side stretching while the other relaxes when the finger swings.

The metacarpophalangeal joint, proximal and distal knuckles have the same configuration. Among them, the proximal phalangeal joint has a high load-bearing capacity and is more frequently used in the human hand. We take it as the object of analysis in GeoGebra, and the analysis is shown in Fig. 8 below.

Analyzing the tendon length $l$ and the moment arm $h$ with $\theta$ as the independent variable:

$$l = \sqrt{L_1^2 + L_2^2 - 2L_1L_2\cos\left(\pi - \theta - arctan\frac{d_2}{R_2} - arctan\frac{d_1}{R_1}\right)} \quad (9)$$

$$h(\theta) = \frac{L_1L_2\sin\alpha}{l}$$

$$= \frac{L_1L_2\sin\alpha}{\sqrt{L_1^2 + L_2^2 - 2L_1L_2\cos\alpha}}$$

$$= \frac{L_1L_2\sin\left(\pi - \theta - arctan\frac{d_2}{R_2} - arctan\frac{d_1}{R_1}\right)}{\sqrt{L_1^2 + L_2^2 - 2L_1L_2\cos\left(\pi - \theta - arctan\frac{d_2}{R_2} - arctan\frac{d_1}{R_1}\right)}} \quad (10)$$

Substitute the data of the proximal joint of the universal finger to plot and analyze. As shown in Fig. 9a, the red line represents the tendon length at the joint and the black line represents the fitted straight line. The tendon length gradually shortens by 16.86 mm over the course of the joint angle from 0° to 110°. It means that at least 20 mm of work needed to be done at the driver end of the straight line. This means that the linear actuator needs to perform work at least a 20mm distance. As shown in Fig. 9b, the red line represents the force arm of the tendon and the black line represents the fitted straight line, which increases over the course of the joint angle from 0° to 90°. The increase in the force arm reduces the stretching force on the tendon, making it easier for the finger to perform the bending maneuver under load and to output a greater force/moment at the end of the finger.

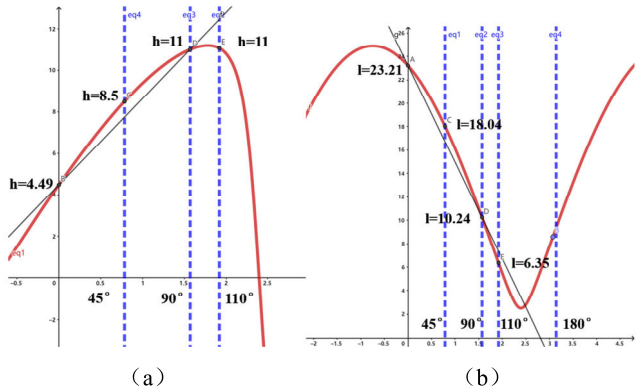

Figure 9. Analysis of tendon lengths and tendon force arm lengths.

### III. DESIGN AND CONSTRUCTION OF THE DRIVE SYSTEM

The integrated solution of the driven system of the CYJ Hand-0 is an important innovation of the overall design. Unlike traditional motor or pneumatic/hydraulic drives, the bionic dexterity hand introduces not only a DC brush motor drive module system to drive the finger flexion, but also an SMA linear drive module system to drive the finger side-swing and straightening movements. They form the SMA-motor hybrid driven module system together.

#### A. Abbreviations and Acronyms

The SMA driven module uses Ni-Ti alloy wires of DYNALLOY from US (detailed parameters as shown in Table III). The SMA material has a shape memory effect, which is a deformability and workability resulting from the martensite-austenite phase change that occurs between low and high temperatures of the material. The method of controlling the module is energized heating. The maximum displacement output is 18 mm and the maximum output force is 784 N. The module construction is shown in Fig. 10.

TABLE III. PARAMETERS OF SMA WIRE

| Parameters | Value |
|---|---|
| Density | 6.45g/cm³ |
| Phase transition temperature | 90°C |
| Specific heat capacity | 837J/(kg·K) |
| Thermal conductivity | 18W/(m·K) |
| Poisson's ratio | 0.33 |
| Coefficient of martensitic thermal expansion | $6.6 \times 10^{-6}$/°C |
| Coefficient of austenitic thermal expansion | $11.0 \times 10^{-6}$/°C |

#### B. Motor driven module

Finger flexion is driven by a screw nut motor (GA12-N20-298, parameters shown in Table IV) drive module. The module drives the tendons to rotate the phalanges relative to the joints so that the fingers are flexed. A single motor driven module has dimensions of 73.25mm in height, 33mm in length, and 25mm in width, with a working stroke of 20mm, a self-weight of 40g, and a rated output force of 400N. The models of the motor and the motor driven module are shown in Fig. 11.

TABLE IV. PARAMETERS OF SMA WIRE

| Parameters | Value |
|---|---|
| Rated voltage | 12 V |
| Rated current | 0.18 A |
| Rated speed | 100 RPM |

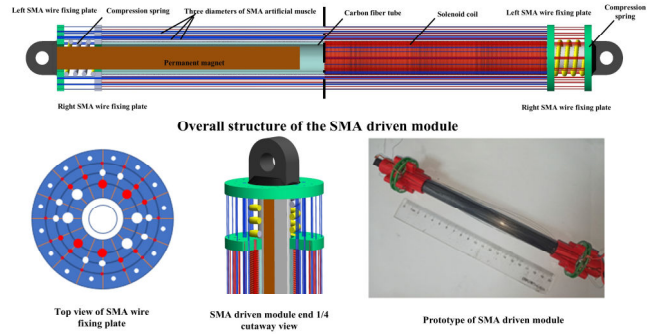

Figure 10. The SMA driven module.

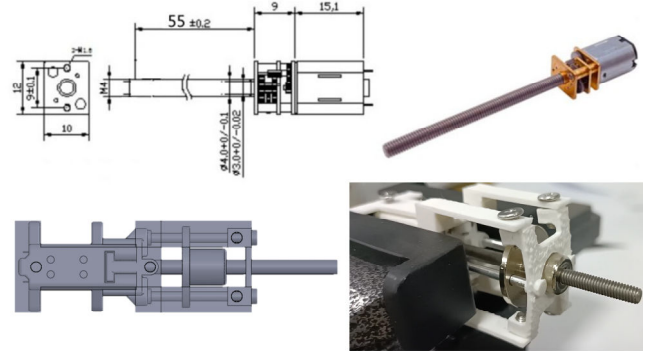

Figure 11. The models of the motor and the motor driven module

| | |
|---|---|
| Rated torque | 1.5 kg · cm |
| Pitch of screw | 0.7 mm |

#### C. The integrated SMA-motor hybrid driven system

In the SMA driven module, SMA wires of larger diameter have a lower response frequency and higher output force. SMA wires of smaller diameter have a higher response frequency and lower output force. Besides, the electromagnetic coil and permanent magnet in the SMA driven module can make up for the shortcomings of the SMA wires. They can form a linear driven module with adjustable size of output force, frequency and fast response speed. The motor driven module features a small but stable output force, self-locking capability, high output displacement accuracy and slow speed (1.17mm/s). As a result, the hybrid driven system constructed by combining the two driven modules has the particularly comprehensive advantages of adjustable response frequency, high output force range and high output displacement accuracy.

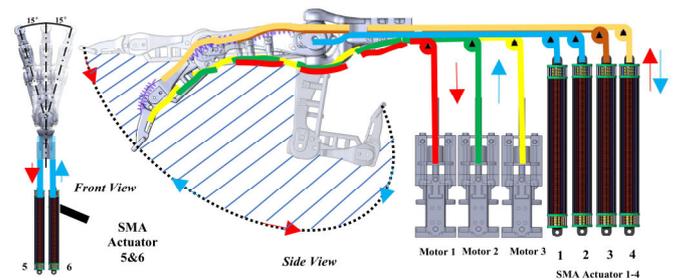

Figure 12 The SMA-motor hybrid driven system.

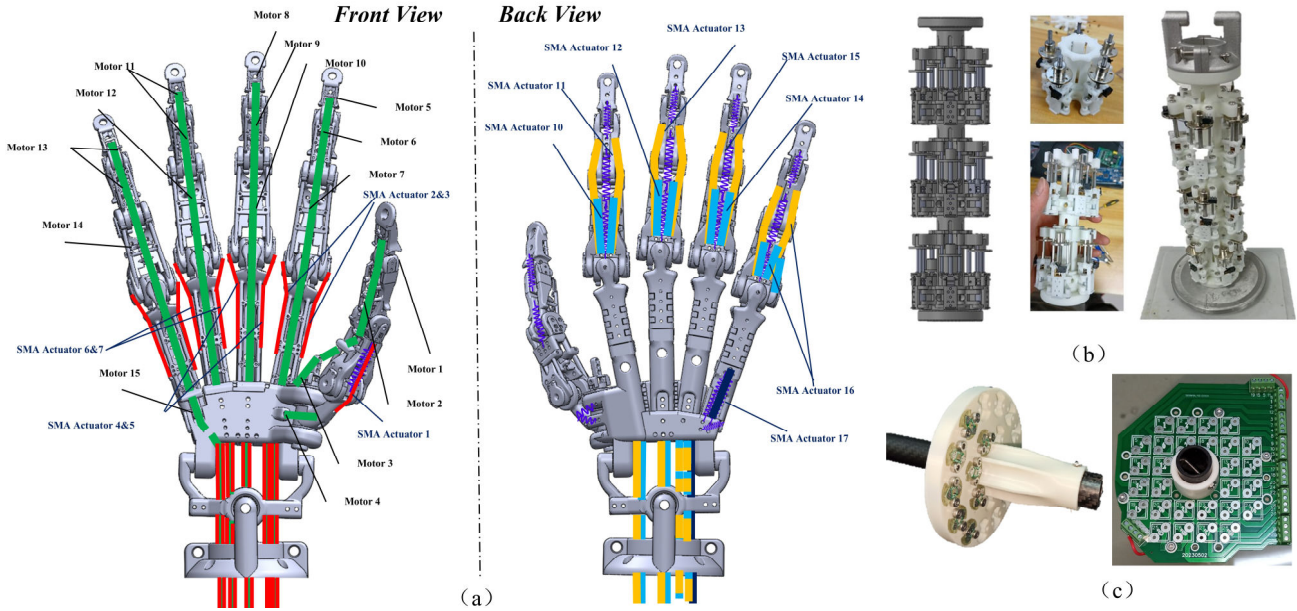

**Figure 13. Driven arrangement and driven system entities of the CYJ Hand-0.**

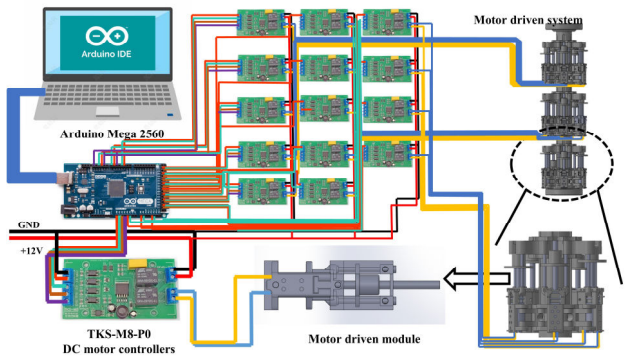

**Figure 14. Motor driven module control system.**

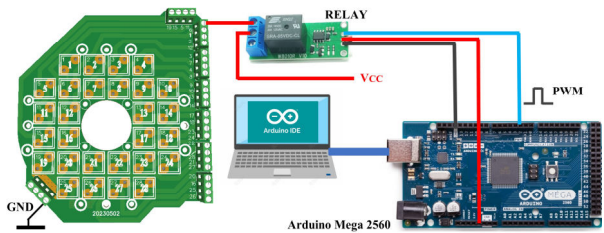

**Figure 15. SMA driven module control system.**

As shown in Fig. 12, the motor driven module drives the finger flexion. With the help of the spring, the SMA linear actuator module drives the finger to straighten or to swing sideways. When the finger is flexed, the motor turns the screw forcibly, causing the nut to drop and the flexor tendon to be stretched. At this time, the SMA linear drive module maintains its original length and the extensor tendon relaxes. As the finger is straightened, the motor drives the screw to reverse the rotation of the nut, causing the flexor tendon to relax. At this point the SMA driven module is energized and heated and the extensor tendon is tensed. The SMA wires and the spring on the back of the phalanx together drive the phalanx to extend until it returns to its original position. After the finger is returned to its original position, the SMA driven module is switched off to dissipate heat and the extensor tendons are relaxed as the SMA wires are restored to their original length.

The CYJ Hand-0 has a total of 21 DOFs. The total number of actuators (TNA) is 32, with a total of 17 SMA driven modules and 15 motor driven modules. The overall driven arrangement is shown in Fig. 13a. In the figure, the motor refers to the end position of the tendon driven by the motor driven module and the SMA actuator refers to the end position of the tendon driven by the SMA driven module.

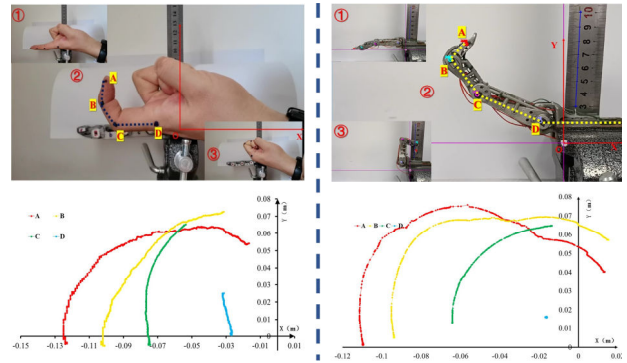

**Figure 16. The CYJ Hand-0 finger flexion experiment.**

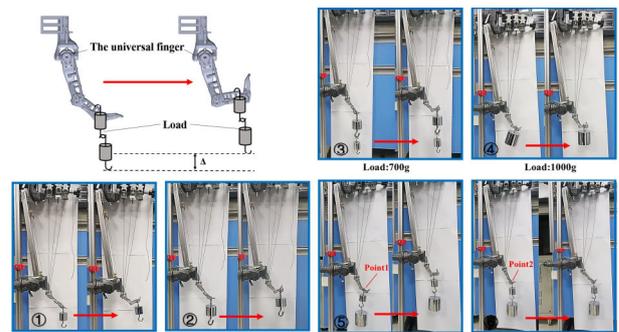

**Figure 17. The CYJ Hand-0 finger tension experiments.**

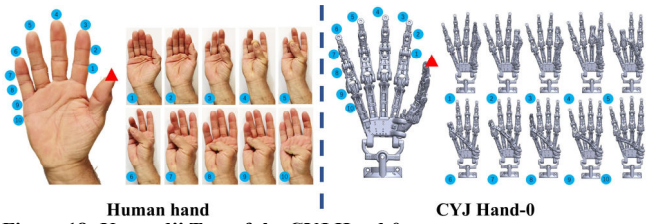

**Figure 18. Kapandji Test of the CYJ Hand-0.**

Numerous external SMA driven modules and motor driven modules must be integrated into the arm attached to the end of the wrist of the bionic dexterous hand to facilitate the transmission of force and displacement through tendons. As shown in Fig.13b, a robotic arm is constructed by arranging 15 motor driven modules in a configuration of five motor driven modules circularly arranged on one layer, stacked over three layers. The robotic arm is 3D printed in white resin with a total weight of 780 g. Its upper part is connected to the wrist of the bionic dexterous hand, and the lower part is matched with the end of UR10. Its hollow structure is used to transfer the tendons. A gap is left between every two motor driven modules to arrange an SMA driven module. 12 SMA driven modules are integrated as shown in Fig. 13c.

### D. Control system

The CYJ Hand-0 is equipped with an electrical motor driven module control system, which is based on the Arduino Mega2560 and 15 Tonkeys TKS-M8-P0 DC motor controllers, as shown in Fig 14. The computer controls the corresponding motor controllers by sending commands to the microcontroller, enabling controllable forward/reverse rotation and start/stop functionality for the different motors.

For the SMA driven module system, a PWM-relay control system as shown in Figure 15 is used—the relays controlling the power supply on/off for each SMA driven module are managed by the high and low level PWM signals output from the Arduino microcontroller.

## IV. EXPERIMENTS

Based on the aforementioned modeling and theoretical analysis, the following sections evaluate the structural feasibility, actuation performance, and overall biomimicry and dexterity of the CYJ Hand-0 through a series of experiments.

### A. Finger experiments

In the CYJ Hand-0 finger flexion experiment, the video processing software Tracker is used for image analysis. Tracker analyses each frame to obtain the time-position information of each joint node, so as to obtain the value of the joint angle at different moments and perform subsequent processing, as shown in Figure 16.

In the finger tension experiments of the CYJ Hand-0, the distal phalanx was selected as the output point for tension measurement, consistent with human usage habits. The load applied to the universal finger was gradually increased from 200 g to 1200 g, and all tests were successfully completed, as shown in Fig. 17. Although the fingertip pulling force did not reach the maximum level of human fingers, the structural load-bearing capacity of a single finger exceeded 4 kg. Notably, during the pulling tests exceeding 1.5 kg, the motor experienced overload and burned out. A single-finger load of 1.2 kg implies that the entire CYJ Hand-0 can support a total load of approximately 8 kg.

### B. Gesture experiments

The first experiment was a CYJ Hand-0 gesture simulation.

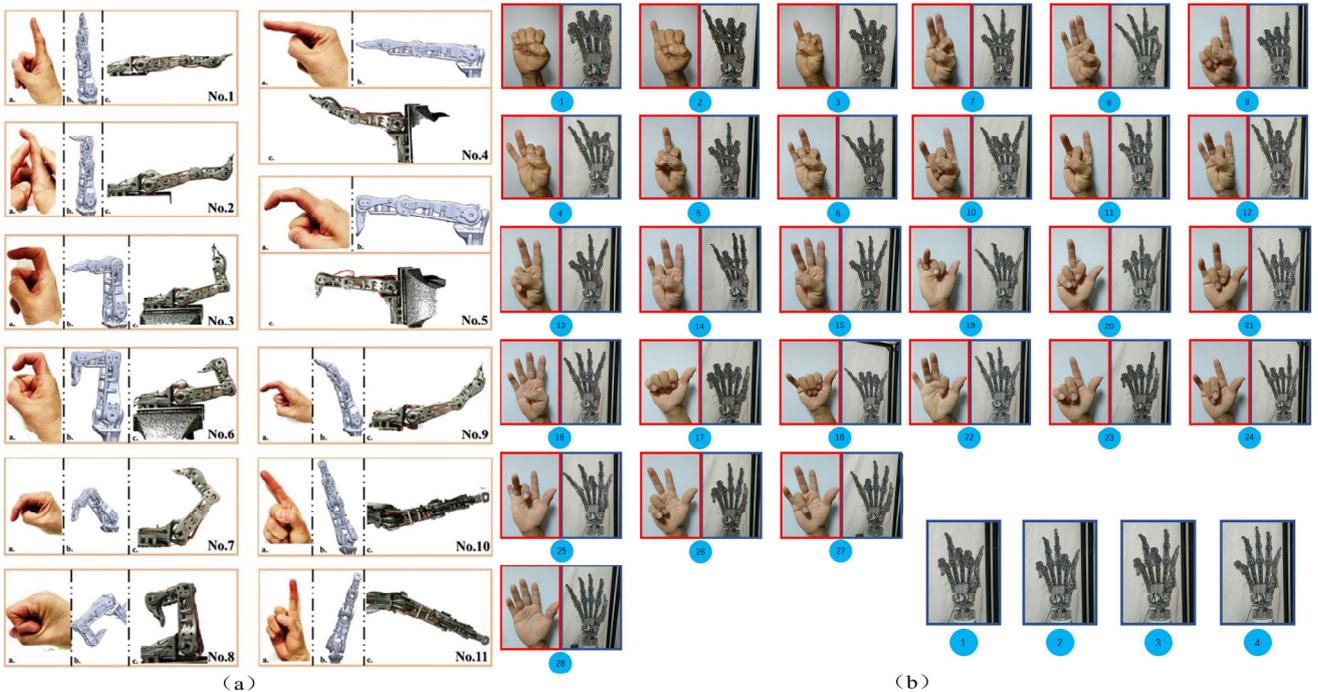

**Figure 19. Gesture experiments of the CYJ Hand-0.**

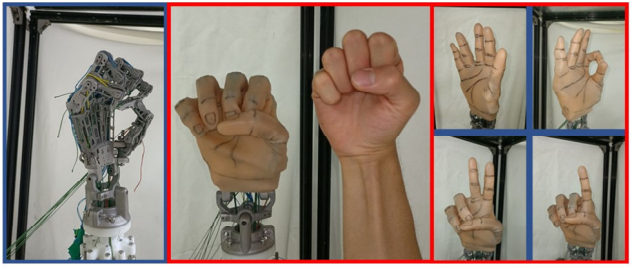

**Figure 20. The CYJ Hand-0 with flesh and skin.**

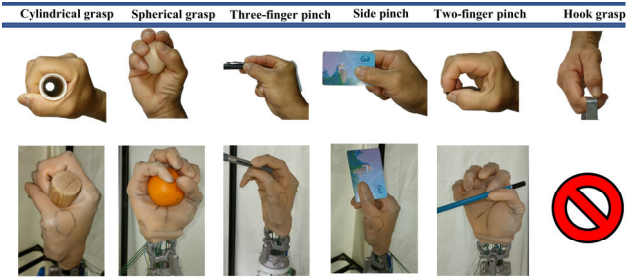

**Figure 21. Schlesinger's 6 grasping postures of the CYJ Hand-0.**

Referring to the test proposed by Ibrahim A. Kapandji in 1985 for measuring the dexterity of the thumb and palm of the human hand (Kapandji Test [14]), whether the tip of the thumb is able to touch the typical part of the hand is taken as the assessment index. As shown in Fig. 18, thanks to its unique structural design, the bionic dexterous hand can perform the same 10 combinations of movements as the human hand.

To further demonstrate the bionic and dexterity of the CYJ Hand-0. In the finger gesture comparison experiment between human hand and the CYJ Hand-0, shown in Fig. 19a, we used 11 gestures commonly used by the index finger, which is more flexible in the human hand. According to the relationship between the movements of the five fingers, there are 28 gestures that the human hand can generally perform without external force, and 4 gestures that are difficult to perform without external force. In the gesture comparison experiment between the human hand and the CYJ Hand-0, as shown in Fig. 19b, the CYJ Hand-0 successfully performed all 32 gestures.

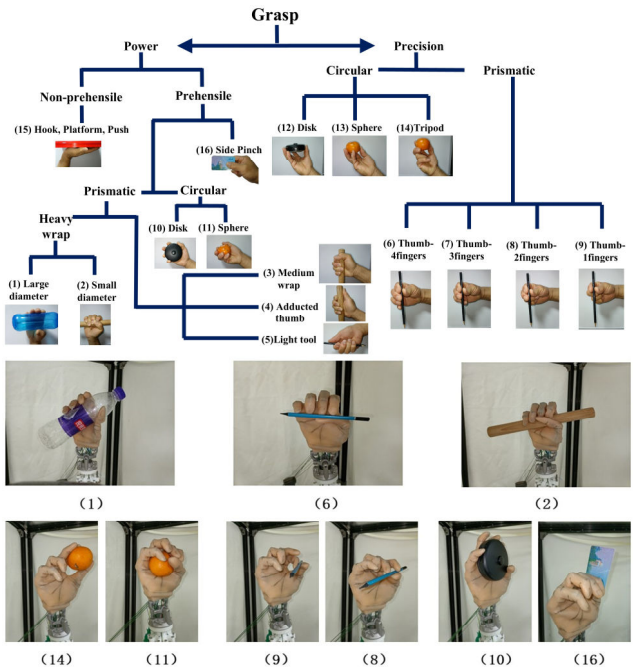

**Figure 22. Cutkosky's 9 grasping postures of the CYJ Hand-0.**

*C. Grasping experiments*

The biggest difference between the human hand and other mechanical grips is its flexibility and versatility. Thanks to the good use of the human hand to develop and use various tools suitable for the human hand, we have embarked on the path of transforming and conquering nature. In order to reflect the function of the CYJ Hand-0, a series of grasping experiments on various objects are carried out below.

According to Schlesinger [15], human hand grasping postures can be divided into six types based on different object shapes and the position used by the human hand: cylindrical grasp, spherical grasp, three-finger pinch, side pinch, two-finger pinch, and hook grasp. Fig. 21 demonstrates five of these postures achieved by the CYJ Hand-0 (the hook grasp was not tested).

Further, the grasping modes of the human hand were categorized into 16 more specific ones according to Cutkosky's classification method [16], as shown in Fig. 22a. The CYJ Hand-0 completes 9 of these postures in the experiment, as shown in Fig. 22b.

In addition to grasping experiments based on traditional classification methods, as shown in Figs. 20 and 21, we also attempted gesture experiments such as grasping a pen (Fig. 22a) and grasping a full-filled water bottle (Fig. 22b), in which the full-filled water bottle was a 550 ml mineral water bottle. It is further verified that the CYJ Hand-0 has the ability to grasp a certain load.

## V. CONCLUSION

In this study, we proposed a high-DOF dexterous hand based on a hybrid SMA–motor drive. This paper describes the overall structural design of the CYJ Hand-0, including key components such as the thumb and metacarpal bone. The hand features 21-DOF, a highly biomimetic and flexible

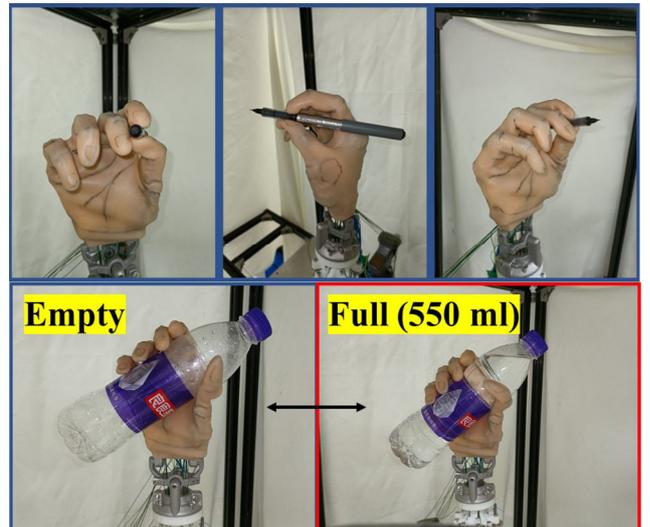

**Figure 23. Gripping experiments with pens and full water bottles.**

structure, and all parts are fabricated using metal 3D printing technology.

This paper also presents the design of the hybrid drive system. The SMA–motor hybrid actuation module we proposed is a key innovation. Comprising motor drive modules and SMA drive modules, the hybrid system offers unique and comprehensive advantages, such as adjustable response frequency, a wide output force range, and high output displacement accuracy. Additionally, we proposed a robotic arm integrated with both motor and SMA drive modules, which is externally mounted at the end of the CYJ Hand-0 to serve as both the power source and support structure. Finally, we developed a basic drive control system based on the Arduino Mega2560 microcontroller.

In future work, we plan to optimize the structural design and actuation scheme of the CYJ Hand-0 based on experimental feedback. We will also improve the sensor system and closed-loop control system to further enhance the application value of the hand.

## References

[1] O. M. Andrychowicz *et al.*, "Learning dexterous in-hand manipulation," *The International Journal of Robotics Research*, vol. 39, no. 1, Art. no. 1, Jan. 2020.

[2] R. Tomovic and G. Boni, "An adaptive artificial hand," *IRE Trans. Autom. Control*, vol. 7, no. 3, pp. 3–10, 1962.

[3] T. Okada, "Computer Control of Multijointed Finger System for Precise Object-Handling," *IEEE Trans. Syst., Man, Cybern.*, vol. 12, no. 3, Art. no. 3, 1982.

[4] C. Loucks, V. Johnson, P. Boissiere, G. Starr, and J. Steele, "Modeling and control of the stanford/JPL hand," in *Proceedings. 1987 IEEE International Conference on Robotics and Automation*, Raleigh, NC, USA: Institute of Electrical and Electronics Engineers, 1987, pp. 573–578.

[5] S. C. Jacobsen, J. E. Wood, D. F. Knutti, and K. B. Biggers, "The UTAH/M.I.T. Dextrous Hand: Work in Progress," *The International Journal of Robotics Research*, vol. 3, no. 4, Art. no. 4, Dec. 1984.

[6] C. S. Lovchik and M. A. Diftler, "The robonaut hand: a dexterous robot hand for space," in *Proceedings 1999 IEEE International Conference on Robotics and Automation (cat. No.99ch36288c)*, 1999, pp. 907–912 vol.2.

[7] M. C. Carrozza, G. Cappiello, S. Micera, B. B. Edin, L. Beccai, and C. Cipriani, "Design of a cybernetic hand for perception and action," *Biol. Cybern.*, vol. 95, no. 6, pp. 629–644, 2006.

[8] Bart *et al.*, "UT hand I: a lock-based underactuated hand prosthesis," *Mech. Mach. Theory*, 2014.

[9] H. Gray, H. Gray, and W. H. Lewis, "Anatomy of the Human Body.," *Postgraduate Medical Journal*, vol. 42, no. 493, Art. no. 493, 1966.

[10] John Lin, Ying Wu, and T. S. Huang, "Modeling the constraints of human hand motion," in *Proceedings Workshop on Human Motion*, Los Alamitos, CA, USA: IEEE Comput. Soc, 2000, pp. 121–126.

[11] L. Y. Chang and N. S. Pollard, "Method for Determining Kinematic Parameters of the *In Vivo* Thumb Carpometacarpal Joint," *IEEE Trans. Biomed. Eng.*, vol. 55, no. 7, Art. no. 7, Jul. 2008.

[12] J. J. Crisco, E. Halilaj, D. C. Moore, T. Patel, A.-P. C. Weiss, and A. L. Ladd, "In Vivo Kinematics of the Trapeziometacarpal Joint During Thumb Extension-Flexion and Abduction-Adduction," *The Journal of Hand Surgery*, vol. 40, no. 2, Art. no. 2, Feb. 2015.

[13] J. H. Buffi, J. J. Crisco, and W. M. Murray, "A method for defining carpometacarpal joint kinematics from three-dimensional rotations of the metacarpal bones captured in vivo using computed tomography," *Journal of Biomechanics*, vol. 46, no. 12, Art. no. 12, Aug. 2013.

[14] A. I. Kapandji, "Clinical test of apposition and counter-apposition of the thumb," *Annales de chirurgie de la main : organe officiel des societes de chirurgie de la main*, vol. 5 1, pp. 67–73, 1986.

[15] G. Schlesinger, "Der mechanische Aufbau der künstlichen Glieder," in *Ersatzglieder und Arbeitshilfen: Für Kriegsbeschädigte und Unfallverletzte*, M. Borchardt, K. Hartmann, Leymann, R. Radike, Schlesinger, and Schwiening, Eds., Berlin, Heidelberg: Springer, 1919, pp. 321–661.

[16] M. R. Cutkosky, "On grasp choice, grasp models, and the design of hands for manufacturing tasks," *IEEE Trans. Robot. Automat.*, vol. 5, no. 3, Art. no. 3, Jun. 1989.